\documentclass[letterpaper, 10 pt, conference, table]{ieeeconf}	
\IEEEoverridecommandlockouts
\pdfoutput=1  
\usepackage{hyperref}
\hypersetup{
    colorlinks,
    citecolor=black,
    filecolor=black,
    linkcolor=black,
    urlcolor=black
}

\makeatletter
\let\NAT@parse\undefined
\makeatother
\usepackage[numbers, sort]{natbib}

\usepackage{times}
\usepackage{graphicx}
\usepackage[usenames, dvipsnames]{color}
\usepackage{subfigure}
\usepackage{bm}
\usepackage{amssymb}
\usepackage{amsfonts}
\usepackage{amsmath}
\usepackage{flushend}
\usepackage{multirow}
\usepackage{url}
\usepackage{comment}
\usepackage{caption}
\usepackage{tikz}
\usetikzlibrary{shapes,arrows,decorations.pathmorphing,backgrounds,positioning,fit,petri}
\usepackage{pgf}
\usepackage{pgfplots}
\usepackage{calc}

\usepackage{array,hhline}

\makeatletter
\newcommand*{\ccol}[1]{%
  \ifdim\dimexpr#1pt/100<.5pt\relax\else\color{white}\fi
  \edef\x{\noexpand\cellcolor[gray]{\strip@pt\dimexpr(100pt-#1pt)/100}}\x
  #1%
}
\newlength{\cellwidth}
\def\jline{\\\hhline{~*{25}{|-}|}}
\makeatother

\newcommand{\wrot}[1]{\w{\rotatebox{90}{#1}}}


\def\secref#1{Sec.~\ref{#1}}
\def\figref#1{Fig.~\ref{#1}}

\def\eqref#1{Eq.~(\ref{#1})}

\definecolor{aisblue}{RGB}{51,102,153}

\renewcommand{\dbltextfloatsep}{0.7em}

\newcommand{\dataset}{Freiburg Groceries Dataset}
\newcommand{\bigSet}{\mathcal{D}_1}
\newcommand{\smallSet}{\mathcal{D}_2}

\newcommand{\githublink}{\url{https://github.com/PhilJd/freiburg_groceries_dataset}}
\newcommand{\aislink}{\url{http://www2.informatik.uni-freiburg.de/\%7Eeitel/freiburg_groceries_dataset.html}}

\begin{document}

\title{\LARGE \bf {The \dataset}}

\author{Philipp Jund, Nichola Abdo, Andreas Eitel, Wolfram Burgard \thanks{All
                authors are with the University of Freiburg, 79110 Freiburg, Germany.  
This work has partly been supported by the German Research Foundation under research unit FOR 1513 (HYBRIS) and grant number EXC 1086.}}

\maketitle
\begin{abstract}
    With the increasing performance of machine learning techniques in the last few years,
    the computer vision and robotics communities have created a large number of
    datasets for benchmarking object recognition tasks. These datasets
    cover a large spectrum of natural images and object categories, making them
    not only useful as a testbed for comparing machine learning approaches, but
    also a great resource for bootstrapping different domain-specific perception and robotic
    systems. One such domain is domestic environments, where an autonomous
    robot has to recognize a large variety of everyday objects such as
    groceries. This is a challenging task due to the large variety of objects
    and products, and where there is great need for \emph{real-world} training
    data that goes beyond product images available online. In this paper, we
    address this issue and present a dataset consisting of 5,000
    images covering 25 different classes of groceries, with at least 97
    images per class. We collected all images from real-world settings at
    different stores and apartments.
    In contrast to existing groceries datasets, our dataset includes a large variety of perspectives, 
    lighting conditions, and degrees of clutter. Overall, our images contain
    thousands of different object instances. It is our hope that machine learning
    and robotics researchers find this dataset of use for training, testing,
    and bootstrapping their approaches. As a baseline classifier to facilitate
    comparison, we re-trained the CaffeNet architecture (an adaptation of the well-known 
    AlexNet~\cite{krizhevsky2012imagenet}) on our dataset and achieved a mean accuracy of 78.9\%. 
    We release this trained model along with the code and data splits we used in our experiments.
    
\end{abstract}

\section{Introduction and Related Work}
\label{sec:intro}

Object recognition is one of the most important and challenging problems in
computer vision. The ability to classify objects plays a crucial role in scene
understanding, and is a key requirement for autonomous robots operating in both
indoor and outdoor environments. Recently, computer vision has witnessed
significant progress, leading to impressive performance in various
detection and recognition
tasks~\cite{he2015delving,taigman2014deepface,ioffe2015batchnorm}.
On the one hand, this is partly due to the recent advancements in
machine learning techniques such as deep learning, fueled by a great interest
from the research community as well as a boost in hardware performance. 
On the other hand, publicly-available datasets have been a
great resource for bootstrapping, testing, and comparing these techniques.

Examples of popular image datasets include ImageNet, CIFAR, COCO and PASCAL,
covering a wide range of categories including people, animals, everyday
objects, and much
more~\cite{imagenet2009deng,krizhevsky2009cifar,everingham2010pascal,lin2014microsoftcoco}.
Other datasets are tailored towards specific domains such as house numbers extracted from Google Street
View~\cite{netzer2011reading}, face recognition~\cite{huang2007labeled}, scene
understanding and place recognition~\cite{zhou2014learning,
silberman2011indoor}, as well as object recognition, manipulation and pose estimation for
robots~\cite{kasper2012kit, calli2015ycbbenchmarking, rennie2016dataset, hinterstoisser2012model}.

One of the challenging domains where object recognition plays a key role is
service robotics. A robot operating in unstructured, domestic environments 
has to recognize everyday objects in order to successfully perform tasks 
like tidying up, fetching objects, or assisting elderly
people. For example, a robot should be able to recognize grocery objects in
order to fetch a can of soda or to predict the preferred shelf
for storing a box of cereals~\cite{srinivasa2010herb, abdo2016organizing}. This is 
not only challenging due to the difficult lighting conditions and occlusions in 
real-world environments, but also due to the large number of everyday objects 
and products that a robot can encounter. 
\begin{figure}[t]
    \includegraphics[width=\columnwidth]{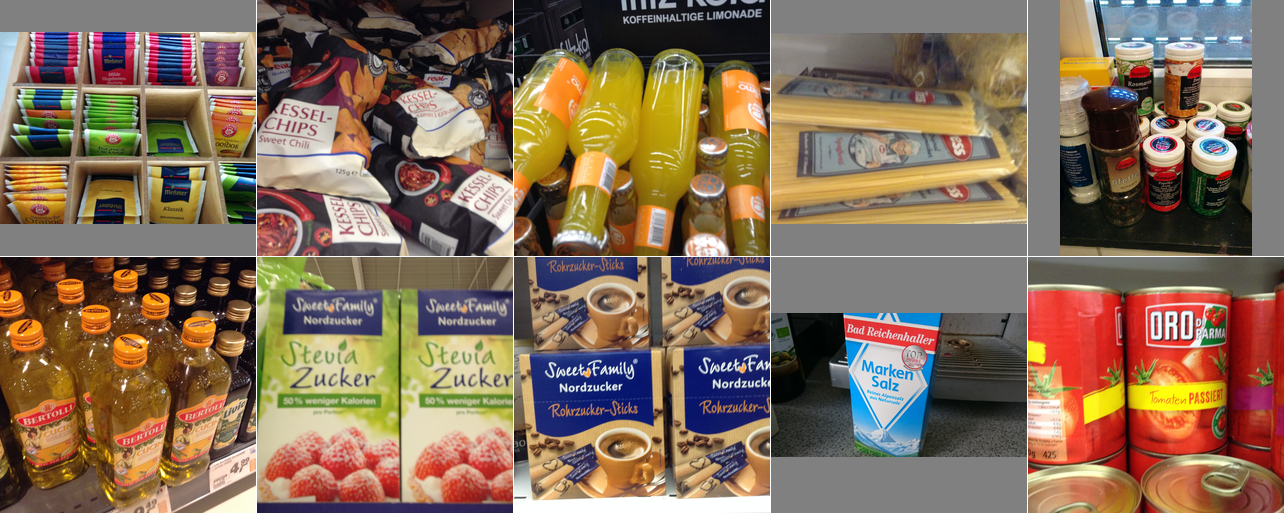}
    \caption{Examples of images from our dataset. Each image contains one or multiple instances of objects 
            belonging to one of 25 classes of groceries. We collected
            these images in real-world settings at different stores and
            apartments.
            We considered a rich variety of perspectives, degree of clutter,
            and lighting conditions. This dataset highlights the challenge 
            of recognizing objects in this domain due to the large variation 
            in shape, color, and appearance of everyday products even within 
            the the same class.
    }
    \label{fig:challengingexamples}
\end{figure}

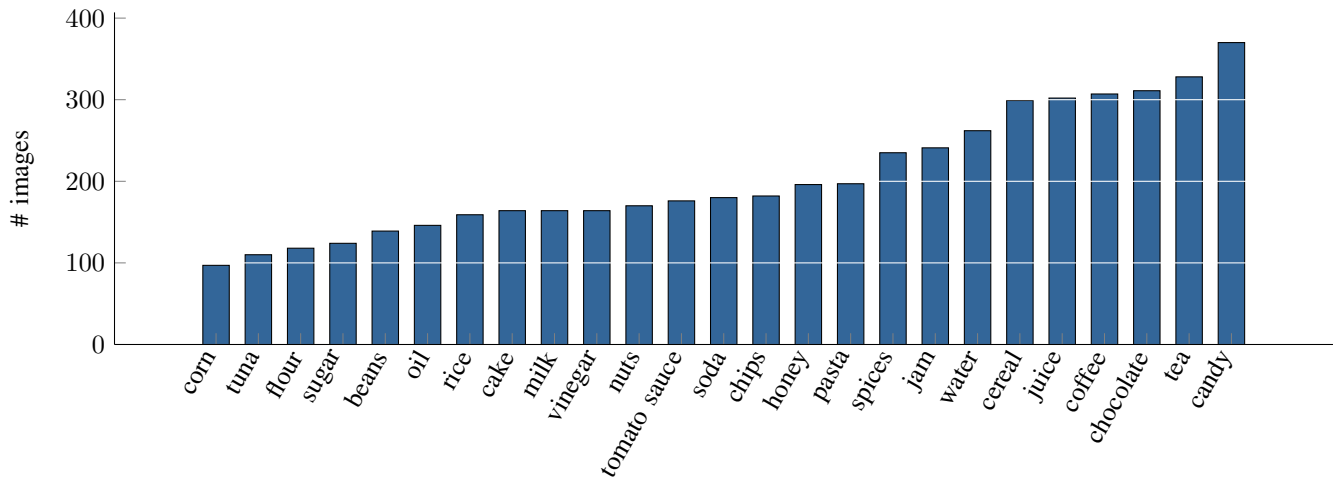
\begin{figure*}
    \begin{tikzpicture}
        \begin{axis}[
            width=\textwidth,
            height=6cm,
            ymin=0,
            ylabel={\# images},
            axis lines*=left,
            axis on top,
            major grid style=white,
            ymajorgrids,
            x tick label style={rotate=60, anchor=east},
            symbolic x coords={corn, tuna, flour, sugar, beans, oil, rice, cake, milk, vinegar, nuts, {tomato sauce}, soda, chips, honey, pasta, spices, jam, water, cereal, juice, coffee, chocolate, tea, candy},
            xtick=data]

                \addplot[ybar,fill=aisblue] coordinates {
                    (corn, 97)
                    (tuna, 110)
                    (flour, 118)
                    (sugar, 124)
                    (beans, 139)
                    (oil, 146)
                    (rice, 159)
                    (cake, 164)
                    (milk, 164)
                    (vinegar, 164)
                    (nuts, 170)
                    ({tomato sauce}, 176)
                    (soda, 180)
                    (chips, 182)
                    (honey, 196)
                    (pasta, 197)
                    (spices, 235)
                    (jam, 241)
                    (water, 262)
                    (cereal, 299)
                    (juice, 302)
                    (coffee, 307)
                    (chocolate, 311)
                    (tea, 328)
                    (candy, 370)
                };
        \end{axis}
    \end{tikzpicture}
\caption{Number of images per class in our dataset.}
\label{fig:numimagesbarplot}
\end{figure*}

Typically, service robotic systems address the problem of object recognition
for different tasks by relying on state-of-the-art perception methods.
Those methods leverage existing object models by extracting hand-designed visual and 3D
descriptors in the environment~\cite{hsiao2010making, alexandre20123d, pauwels_simtrack_2015} or by learning new feature representations from raw sensor data~\cite{bo_iser12, eitel15iros}. Others rely on an ensemble of perception techniques and sources of information
including text, inverse image search, cloud data, or images downloaded from online stores
to categorize objects and reason about their relevance for different
tasks~\cite{irosws11germandeli, Kaiser2014, beetz2015robosherlock, kehoe2013ICRA}.

However, leveraging the full potential of machine learning approaches to address 
problems such as recognizing groceries and food items remains, to a large
extent, unrealized. One of the main reasons for that is the lack of training
data for this domain. In this paper, we address this issue and present the
\dataset, a rich collection of 5000 images of grocery products (available in
German stores) and covering 25 common categories. Our motivation for this is twofold: \emph{i)} to
help bootstrap perception systems tailored for domestic robots and assistive 
technologies, and to \emph{ii)} provide a challenging benchmark for testing and
comparing object recognition techniques.

While there exist several datasets containing groceries, they are typically
limited with respect to the view points or variety of instances. 
For example, sources such as the OpenFoodFacts dataset or images available on
the websites of grocery stores typically consider one or two views of
each item~\cite{gigandet2012openfoodfacts}. Other datasets include multiple
view points of each product but consider only a small number of objects. 
An example of this is the GroZi-120 dataset that contains 120 grocery products
under perfect and real lighting conditions~\cite{merler2007insitu}. Another
example is the RGB-D dataset covering 300 object instances in a controlled environment, of which only 
a few are grocery items~\cite{lai2011rgbwashington}. The CMU dataset,
introduced by~\citeauthor{hsiao2010making},
considers multiple viewpoints of 10 different household objects~\cite{hsiao2010making}. Moreover, the BigBIRD dataset contains 3D models and images of 100 instances in a controlled environment~\cite{singh2014bigbird}. 

In contrast to these datasets, the \dataset~considers challenging real-world
scenes as depicted in \figref{fig:challengingexamples}. This includes difficult
lighting conditions with reflections and shadows, as well as different degrees
of clutter ranging from individual objects to packed shelves. Additionally, we consider a large number of instances that cover a
rich variety of brands and package designs.

To demonstrate the applicability of existing machine learning techniques to
tackle the challenging problem of recognizing everyday grocery items, we
trained a convolutional neural network as a baseline classifier on five splits of
our dataset, achieving a classification accuracy of 78.9\%. Along with the
dataset, we provide the code and data splits we used in these experiments.
Finally, whereas each image in the main dataset contains objects belonging to
one class, we include an additional set of 37 cluttered scenes, each containing
several object classes. We constructed these scenes at our lab to emulate real-world 
storage shelves. We present qualitative examples in this paper that demonstrate using our classifier to recognize patches extracted from such images.

\section{The \dataset}
\label{sec:dataset}
\begin{figure*}[t]
    \begin{tikzpicture}
     \newcommand{\imgnode}{\node[anchor=south west,inner sep=0]}
     \newcommand{\pastadir}{figures/positive_examples/PASTA}
     \newcommand{\candydir}{figures/positive_examples/CANDY}
    \imgnode (1) at (0, 0.0) {{\includegraphics[width=0.16\textwidth]{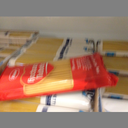}}};
    \imgnode (2) at (1/6*\textwidth, 0) {{\includegraphics[width=0.16\textwidth]{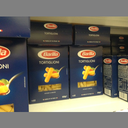}}};
    \imgnode (3) at (2/6*\textwidth, 0) {{\includegraphics[width=0.16\textwidth]{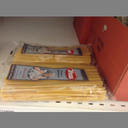}}};
    \imgnode (4) at (3/6*\textwidth, 0) {{\includegraphics[width=0.16\textwidth]{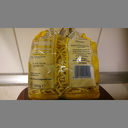}}};
    \imgnode (5) at (4/6*\textwidth, 0) {{\includegraphics[width=0.16\textwidth]{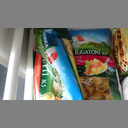}}};
    \imgnode (6) at (5/6*\textwidth, 0) {{\includegraphics[width=0.16\textwidth]{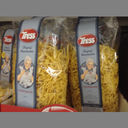}}};
    \imgnode (7) at (0, 3) {{\includegraphics[width=0.16\textwidth]{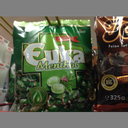}}};
    \imgnode (8) at (1/6*\textwidth, 3) {{\includegraphics[width=0.16\textwidth]{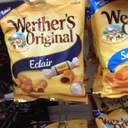}}};
    \imgnode (9) at (2/6*\textwidth, 3) {{\includegraphics[width=0.16\textwidth]{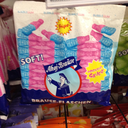}}};
    \imgnode (10) at (3/6*\textwidth, 3) {{\includegraphics[width=0.16\textwidth]{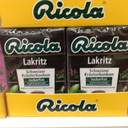}}};
    \imgnode (11) at (4/6*\textwidth, 3) {{\includegraphics[width=0.16\textwidth]{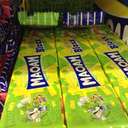}}};
    \imgnode (12) at (5/6*\textwidth, 3) {{\includegraphics[width=0.16\textwidth]{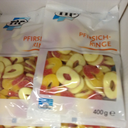}}};
    \end{tikzpicture}
    \caption{Example test images for candy and pasta taken from the first
    split in our experiments. All images were correctly classified in this
    case. The classifier is able to handle large variations in color,
    shape, perspective, and degree of clutter.}
    \label{fig:positiveexamples}
\end{figure*}

The main bulk of the \dataset~consists of 4947 images of 25 grocery classes,
with 97 to 370 images per class.~\figref{fig:numimagesbarplot} shows an
overview of the number of images per class. We considered common categories of
groceries that exist in most homes such as pasta, spices, coffee, etc. 
We recorded this set of images, which we denote by $\bigSet$, using four
different smartphone cameras at various stores (as well as some apartments and offices) in Freiburg, Germany. The images vary in the degree of clutter and real-world lighting conditions, ranging from well-lit stores to kitchen
cupboards. Each image in $\bigSet$ contains one or multiple instances of 
\emph{one} of the 25 classes. Moreover, for each class, we considered a rich
variety of brands, flavors, and packaging designs. We processed all images in
$\bigSet$ by down-scaling them to a size of 256$\times$256 pixels. Due to the
different aspect ratios of the cameras we used, we padded the images with gray
borders as needed.~\figref{fig:imageoverview} shows example images for each class.

Moreover, the \dataset~includes an additional, smaller set $\smallSet$ with 74 images of 37 cluttered scenes, each containing objects belonging to \emph{multiple}
classes. We constructed these scenes at our lab and recorded them using a
Kinect v2 camera \cite{libfreenect2, iai_kinect2}.
For each scene, we provide data from two different camera perspectives, which includes a 1920$\times$1080
RGB image, the corresponding depth image and a point cloud of the scene. 
We created these scenes to emulate real-world clutter and to provide a challenging benchmark with multiple object categories per image. We provide a ``coarse'' labeling of images in $\smallSet$ in
terms of which classes exist in each scene.

We make the dataset available on this website: \aislink. There, we also include the trained classifier model we used in our experiments (see~\secref{sec:classifier}). Additionally, we provide the code for
reproducing our experimental results on github: \githublink.

\section{Object Recognition Using\\a Convolutional Neural Network}
\label{sec:classifier}

To demonstrate the use of our dataset and provide a baseline for future
comparison, we trained a deep neural network classifier using the images in $\bigSet$. 
We adopted the CaffeNet architecture~\cite{jia2014caffe}, a slightly altered version of the
AlexNet~\cite{krizhevsky2012imagenet}. We trained this model, which consists 
of five convolution layers and three fully connected layers, using the Caffe 
framework~\cite{jia2014caffe}. 
We initialized the weights of the model with those of the
pre-trained CaffeNet, and fine-tuned the weights of the three fully-connected layers.

\begin{figure}[t]
    \begin{tikzpicture}
    
     \newcommand{\imgnode}{\node[anchor=south west,inner sep=0]}

    \imgnode (1) at (0, 0.0) {{\includegraphics[width=0.31\columnwidth]{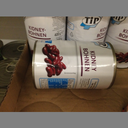}}};
    \node (1x) [below=of 1, align=left, yshift=20pt] {Class: Beans \\ Predicted: Sugar};

    \imgnode (2) at (0.333\columnwidth, 0.0) {{\includegraphics[width=0.31\columnwidth]{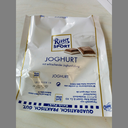}}};
    \node (2x) [below=of 2, align=left, yshift=20pt] {Class: Chocolate \\ Predicted: Flour};

    \imgnode (3) at (0.666\columnwidth, 0.0) {{\includegraphics[width=0.31\columnwidth]{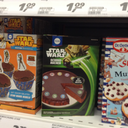}}};
    \node (3x) [below=of 3, align=left, yshift=20pt] {Class: Cake \\ Predicted: Chocolate};

    \imgnode (4) at (0, -4.5) {{\includegraphics[width=0.31\columnwidth]{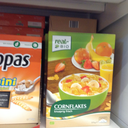}}};
    \node (4x) [below=of 4, align=left, yshift=20pt] {Class: Cereal \\ Predicted: Juice};

    \imgnode (5) at (0.333\columnwidth, -4.5) {{\includegraphics[width=0.31\columnwidth]{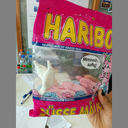}}};
    \node (5x) [below=of 5, align=left, yshift=20pt] {Class: Candy \\ Predicted: Coffee};

    \imgnode (6) at (0.666\columnwidth, -4.5) {{\includegraphics[width=0.31\columnwidth]{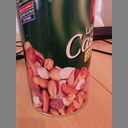}}};
    \node (6x) [below=of 6, align=left, yshift=20pt] {Class: Nuts \\ Predicted: Beans};

    \end{tikzpicture}
    \caption{Examples of misclassification that highlight some of the
    challenges in our dataset, e.g., cereal boxes with drawings of
    fruits on them are sometimes confused with juice.
    }
    \label{fig:misclassified}
\end{figure}

We partitioned the images into five equally-sized splits, with the images of each
class uniformly distributed over the splits. We used each split as a test set
once and trained on the remaining data. In each case, we balanced the 
training data across all classes by duplicating images from classes with fewer
images.
We trained all models for 10,000 iterations and always used the last model for evaluation.

We achieved a mean accuracy of 78.9\% (with a standard deviation of 0.5\%)
over all splits.~\figref{fig:positiveexamples} shows examples of correctly
classified images of different candy and pasta packages. The neural network is 
able to recognize the categories in these images despite large variations in
appearance, perspectives, lighting conditions, and number of objects in each
image. On the other hand,~\figref{fig:misclassified} shows examples of
misclassified images. For example, we found products with white, plain
packagings to be particularly challenging, often mis-classified as flour.
Another source of difficulty is products with ``misleading'' designs, e.g.,
pictures of fruit (typically found on juice cartons) on cereal boxes.

\figref{fig:confmat} depicts the confusion matrix averaged over the five splits. 
The network performs particularly well for classes such as water, jam, and
juice (88.1\%-93.2\%), while on the other hand it has difficulties correctly recognizing objects from the class flour (59.9\%).
We provide the data splits and code needed to reproduce our results, along with
the a Caffe model trained on all images of $\bigSet$, on the web pages mentioned in~\secref{sec:dataset}.

\begin{figure}[t]
    \begin{tikzpicture}
    
     \newcommand{\imgnode}{\node[anchor=south west,inner sep=0]}
     \newcommand{\dir}{figures/cluttered_scene/}

    \imgnode (1) at (0, 0.0) {{\includegraphics[width=0.69\columnwidth]{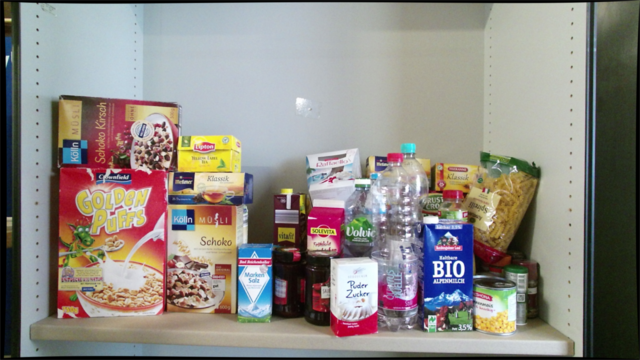}}};
    \node (1x) [below=of 1, align=center, yshift=25pt] {(a)};
    \imgnode (3) at (0.7\columnwidth, 0) {{\includegraphics[width=0.3\columnwidth]{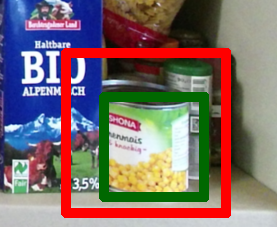}}};
    \node (3x) [below=of 3, align=center, yshift=25pt] {(b)};
    \imgnode (2) at (0, -5.8) {{\includegraphics[width=\columnwidth]{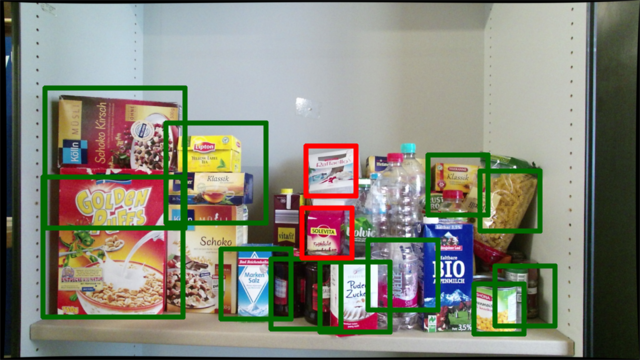}}};
    \node (2x) [below=of 2, align=center, yshift=25pt] {(c)};

    \end{tikzpicture}
    \caption{We used a classifier trained on $\bigSet$ to recognize manually-selected patches from dataset $\smallSet$. Patches with a green border indicate a correct classification whereas those with a red
    border indicate a misclassification. We rescaled each patch
    before passing it through the network.
    (a)~depicts the complete scene. (b)~shows an example of the sensitivity of
    the network to changes in patch size. (c)~shows classification results for some manually extracted patches.}
    \label{fig:cluttered}
\end{figure}

Finally, we also performed a qualitative test where we used a model trained 
on images in $\bigSet$ to classify patches extracted from images 
in $\smallSet$ (in which each image contains multiple object classes).
\figref{fig:cluttered} shows an example for classifying different 
manually-selected image patches. Despite a sensitivity to patch
size, this shows 
the potential for using $\bigSet$, which only includes one class per image, to
recognize objects in cluttered scenes where this assumption does not hold.
An extensive evaluation on such scenes is outside the scope of this paper, which we leave to future work.

\section{Conclusion}
In this paper, we introduced the~\dataset, a novel dataset targeted at the
recognition of groceries. Our dataset includes ca 5000 labeled images, organized in 25 
classes of products, which we recorded in several stores
and apartments in Germany. Our images cover a wide range of real-world
conditions including different viewpoints, lighting conditions, and degrees of clutter. 
Moreover, we provide images and point clouds for a set of 37 cluttered scenes, each consisting of objects from multiple classes. To facilitate comparison, we provide results averaged
over five train/test splits using a standard deep network architecture, which
achieved a mean accuracy of 78.9\% over all classes. We believe that the
\dataset~represents an interesting and challenging benchmark to evaluate 
state-of-the-art object recognition techniques such as deep neural networks.
Moreover, we believe that this real-world training data is a valuable resource for
accelerating a variety of service robot applications and assistive systems
where the ability to recognize everyday objects plays a key role.

\footnotesize
\bibliographystyle{abbrvnat}
\bibliography{refs}

\newpage
\begin{figure*}[t]
\scriptsize
\setlength\tabcolsep{0.07cm}
\renewcommand{\arraystretch}{1.6}
\begin{tabular}{l*{26}{|>{\centering\arraybackslash}m{0.7\cellwidth}}|}

\noalign{\gdef\w#1{\multicolumn{1}{c}{{\small#1}}}}%
\w{}                    & \wrot{beans}  & \wrot{cake} & \wrot{candy}	& \wrot{cereal}	 & \wrot{chips}  & \wrot{chocolate} 	& \wrot{coffee}  & \wrot{corn} & \wrot{flour} 	& \wrot{honey}  & \wrot{jam}  & \wrot{juice} & \wrot{milk} & \wrot{nuts}	& \wrot{oil} 	& \wrot{pasta} 	 & \wrot{rice} 	 	 & \wrot{soda}	 & \wrot{spices}	   	 & \wrot{sugar}   & \wrot{tea}  & \wrot{tomato sauce} & \wrot{tuna}	& \wrot{vinegar} 	& \wrot{water} 		 \jline
{\small beans} 			& \ccol{78.1}   & \ccol{0.5}  & \ccol{0.5} 	    & \ccol{1.9} 	 & \ccol{0.0} 	 & \ccol{1.5} 	 		& \ccol{0.0} 	 & \ccol{3.1}  & \ccol{0.0} 	& \ccol{1.8} 	& \ccol{2.8}  & \ccol{0.8} 	 & \ccol{0.8}  & \ccol{0.8} 	& \ccol{0.5}	& \ccol{0.8} 	 & \ccol{0.8} 		 & \ccol{0.0} 	 & \ccol{1.0} 	 	 	 & \ccol{0.8} 	  & \ccol{0.8}  & \ccol{1.1} 	 		& \ccol{0.0 } 	& \ccol{0.0} 	 	& \ccol{0.8} 	 \jline
{\small cake} 		 	& \ccol{0.0} 	& \ccol{82.5} & \ccol{0.6} 	    & \ccol{5.2} 	 & \ccol{0.9} 	 & \ccol{2.9} 	 		& \ccol{1.1} 	 & \ccol{0.6}  & \ccol{0.0} 	& \ccol{0.0} 	& \ccol{0.0}  & \ccol{0.0} 	 & \ccol{0.4}  & \ccol{0.4} 	& \ccol{0.0}	& \ccol{1.9} 	 & \ccol{0.0} 		 & \ccol{0.0} 	 & \ccol{0.0} 	 	  	 & \ccol{0.0} 	  & \ccol{1.8}  & \ccol{0.6} 	 		& \ccol{0.6 } 	& \ccol{0.0} 	 	& \ccol{0.0} 	 \jline
{\small candy} 		 	& \ccol{0.0} 	& \ccol{0.5}  & \ccol{82.9} 	& \ccol{1.1} 	 & \ccol{1.7} 	 & \ccol{1.9} 	 		& \ccol{0.8} 	 & \ccol{0.0}  & \ccol{0.2} 	& \ccol{0.2} 	& \ccol{0.5}  & \ccol{1.0} 	 & \ccol{1.1}  & \ccol{1.9} 	& \ccol{0.0}	& \ccol{0.2} 	 & \ccol{0.2} 		 & \ccol{0.0} 	 & \ccol{0.0} 	 	 	 & \ccol{0.2} 	  & \ccol{1.9}  & \ccol{1.6} 	 		& \ccol{0.6 } 	& \ccol{0.0} 	 	& \ccol{0.8} 	 \jline
{\small cereal}  		& \ccol{0.3} 	& \ccol{4.8}  & \ccol{2.9} 	    & \ccol{78.2} 	 & \ccol{0.3} 	 & \ccol{1.0} 	 		& \ccol{0.6} 	 & \ccol{0.3}  & \ccol{0.7} 	& \ccol{0.0} 	& \ccol{0.0}  & \ccol{1.1} 	 & \ccol{0.6}  & \ccol{1.8} 	& \ccol{0.4}	& \ccol{1.3} 	 & \ccol{0.7} 		 & \ccol{0.0} 	 & \ccol{1.0} 	 	  	 & \ccol{0.6} 	  & \ccol{1.4}  & \ccol{0.4} 	 		& \ccol{0.3 } 	& \ccol{0.0} 	 	& \ccol{0.4} 	 \jline
{\small chips} 			& \ccol{2.2} 	& \ccol{1.6}  & \ccol{7.1} 	    & \ccol{1.1} 	 & \ccol{70.1}   & \ccol{1.6} 	 		& \ccol{1.1} 	 & \ccol{0.5}  & \ccol{0.5} 	& \ccol{0.0} 	& \ccol{0.0}  & \ccol{6.0} 	 & \ccol{0.0}  & \ccol{4.4} 	& \ccol{0.0}	& \ccol{1.6} 	 & \ccol{0.0} 		 & \ccol{0.0} 	 & \ccol{0.0} 	 	 	 & \ccol{0.5} 	  & \ccol{0.5}  & \ccol{0.5} 	 		& \ccol{0.0 } 	& \ccol{0.0} 	 	& \ccol{0.0} 	 \jline
{\small chocolate} 		& \ccol{0.0} 	& \ccol{3.1}  & \ccol{2.0} 	    & \ccol{4.0} 	 & \ccol{1.0} 	 & \ccol{70.4} 	 	    & \ccol{4.7} 	 & \ccol{0.0}  & \ccol{1.3} 	& \ccol{0.9} 	& \ccol{0.3}  & \ccol{0.6} 	 & \ccol{0.9}  & \ccol{3.3} 	& \ccol{0.0} 	& \ccol{0.6} 	 & \ccol{0.0} 		 & \ccol{0.7} 	 & \ccol{0.0} 	 	 	 & \ccol{0.2} 	  & \ccol{4.8}  & \ccol{0.3} 	 		& \ccol{0.0 } 	& \ccol{0.0} 	 	& \ccol{0.0} 	 \jline
{\small coffee} 		& \ccol{0.0} 	& \ccol{0.3}  & \ccol{0.3} 	    & \ccol{1.3} 	 & \ccol{1.7} 	 & \ccol{5.1} 	 		& \ccol{71.9} 	 & \ccol{0.0}  & \ccol{0.7} 	& \ccol{1.0} 	& \ccol{3.9}  & \ccol{1.3} 	 & \ccol{1.0}  & \ccol{0.9} 	& \ccol{1.5}	& \ccol{0.3} 	 & \ccol{0.0} 		 & \ccol{0.6} 	 & \ccol{2.4} 	 	 	 & \ccol{0.3} 	  & \ccol{2.2}  & \ccol{0.3} 	 		& \ccol{0.9 } 	& \ccol{0.0} 	 	& \ccol{1.3} 	 \jline
{\small corn} 			& \ccol{0.9} 	& \ccol{0.0}  & \ccol{0.0} 	    & \ccol{0.0} 	 & \ccol{0.0} 	 & \ccol{0.0} 	 		& \ccol{0.9} 	 & \ccol{88.3} & \ccol{0.0} 	& \ccol{1.9} 	& \ccol{0.0}  & \ccol{3.8} 	 & \ccol{0.0}  & \ccol{0.0} 	& \ccol{0.0}	& \ccol{0.0} 	 & \ccol{0.0} 		 & \ccol{0.0} 	 & \ccol{1.0} 	 	 	 & \ccol{0.0} 	  & \ccol{0.9}  & \ccol{1.0} 	 		& \ccol{0.0 } 	& \ccol{0.0} 	 	& \ccol{1.0} 	 \jline
{\small flour} 			& \ccol{2.0} 	& \ccol{1.7}  & \ccol{3.2} 	    & \ccol{5.2} 	 & \ccol{0.0} 	 & \ccol{0.8} 	 		& \ccol{0.9} 	 & \ccol{0.8}  & \ccol{59.9}    & \ccol{0.8} 	& \ccol{0.0}  & \ccol{1.7} 	 & \ccol{3.8}  & \ccol{0.8} 	& \ccol{0.8}	& \ccol{6.4} 	 & \ccol{0.8} 		 & \ccol{0.0} 	 & \ccol{0.9} 	 	 	 & \ccol{2.5} 	  & \ccol{5.4}  & \ccol{0.0} 	 		& \ccol{0.0 } 	& \ccol{0.0} 	 	& \ccol{0.8} 	 \jline
{\small honey} 			& \ccol{0.0} 	& \ccol{0.4}  & \ccol{0.4} 	    & \ccol{0.4} 	 & \ccol{1.1} 	 & \ccol{1.0} 	 		& \ccol{2.3} 	 & \ccol{0.0}  & \ccol{0.0} 	& \ccol{73.6} 	& \ccol{9.2}  & \ccol{2.1} 	 & \ccol{0.0}  & \ccol{1.1} 	& \ccol{0.5}	& \ccol{0.0} 	 & \ccol{0.0} 		 & \ccol{1.0} 	 & \ccol{0.4} 	 	 	 & \ccol{0.0} 	  & \ccol{2.5}  & \ccol{1.1} 	 		& \ccol{0.0 } 	& \ccol{1.0} 	 	& \ccol{1.0} 	 \jline
{\small jam} 	 		& \ccol{0.0} 	& \ccol{0.0}  & \ccol{0.4} 	    & \ccol{0.0} 	 & \ccol{0.3} 	 & \ccol{0.0} 	 		& \ccol{2.5} 	 & \ccol{0.0}  & \ccol{0.0} 	& \ccol{0.8} 	& \ccol{91.4} & \ccol{0.0} 	 & \ccol{0.0}  & \ccol{0.4} 	& \ccol{0.0}	& \ccol{0.0} 	 & \ccol{0.0} 		 & \ccol{0.4} 	 & \ccol{0.7} 	  	 	 & \ccol{0.0} 	  & \ccol{0.7}  & \ccol{1.5} 	 		& \ccol{0.4 } 	& \ccol{0.0} 	 	& \ccol{0.0} 	 \jline
{\small juice}			& \ccol{0.0} 	& \ccol{0.3}  & \ccol{1.0} 	    & \ccol{0.0} 	 & \ccol{0.0} 	 & \ccol{0.7} 	 		& \ccol{0.6} 	 & \ccol{0.0}  & \ccol{0.0} 	& \ccol{0.0} 	& \ccol{0.6}  & \ccol{88.1}  & \ccol{0.4}  & \ccol{0.0} 	& \ccol{0.6}	& \ccol{0.0} 	 & \ccol{0.0} 		 & \ccol{0.6} 	 & \ccol{0.6} 	 	 	 & \ccol{0.0} 	  & \ccol{1.2}  & \ccol{2.0} 	 		& \ccol{0.0 } 	& \ccol{2.0} 	 	& \ccol{0.7} 	 \jline
{\small milk} 			& \ccol{0.0} 	& \ccol{0.0}  & \ccol{0.0} 	    & \ccol{1.0} 	 & \ccol{0.0} 	 & \ccol{0.5} 	 		& \ccol{3.3} 	 & \ccol{0.0}  & \ccol{0.0} 	& \ccol{0.0} 	& \ccol{0.0}  & \ccol{5.7} 	 & \ccol{81.7} & \ccol{0.0} 	& \ccol{0.0}	& \ccol{0.0} 	 & \ccol{0.7} 	 	 & \ccol{0.6} 	 & \ccol{0.5} 	 	 	 & \ccol{1.7} 	  & \ccol{2.6}  & \ccol{1.2} 	 		& \ccol{0.0 } 	& \ccol{0.0} 	 	& \ccol{0.0} 	 \jline
{\small nuts} 			& \ccol{2.2} 	& \ccol{1.2}  & \ccol{2.9} 	    & \ccol{4.1} 	 & \ccol{4.4} 	 & \ccol{8.2} 	 		& \ccol{1.0} 	 & \ccol{0.5}  & \ccol{0.0} 	& \ccol{1.2} 	& \ccol{1.0}  & \ccol{0.4} 	 & \ccol{0.0}  & \ccol{67.7}	& \ccol{0.0}	& \ccol{1.2} 	 & \ccol{0.4} 		 & \ccol{0.0} 	 & \ccol{0.0} 	  	 	 & \ccol{0.0} 	  & \ccol{0.4}  & \ccol{0.9} 	 		& \ccol{0.0 } 	& \ccol{0.0} 	 	& \ccol{1.0} 	 \jline
{\small oil }			& \ccol{0.6} 	& \ccol{0.0}  & \ccol{0.0} 	    & \ccol{0.5} 	 & \ccol{0.8} 	 & \ccol{0.0} 	 		& \ccol{0.0} 	 & \ccol{0.0}  & \ccol{0.6} 	& \ccol{0.0} 	& \ccol{1.3}  & \ccol{4.1} 	 & \ccol{0.6}  & \ccol{0.0} 	& \ccol{78.0}	& \ccol{0.0} 	 & \ccol{0.0} 		 & \ccol{1.6} 	 & \ccol{0.5} 	 	 	 & \ccol{0.0} 	  & \ccol{1.4}  & \ccol{0.0} 	 		& \ccol{0.0 } 	& \ccol{9.3} 	 	& \ccol{0.0} 	 \jline
{\small pasta} 			& \ccol{0.5} 	& \ccol{0.5}  & \ccol{0.5} 	    & \ccol{4.0} 	 & \ccol{1.8} 	 & \ccol{3.4} 	 		& \ccol{3.1} 	 & \ccol{0.0}  & \ccol{1.6} 	& \ccol{0.0} 	& \ccol{1.2}  & \ccol{0.5} 	 & \ccol{0.0}  & \ccol{0.0} 	& \ccol{1.1}	& \ccol{76.6} 	 & \ccol{1.1} 		 & \ccol{0.0} 	 & \ccol{1.0} 	 	 	 & \ccol{0.0} 	  & \ccol{2.3}  & \ccol{0.0} 	 		& \ccol{0.0 } 	& \ccol{0.0} 	 	& \ccol{0.0} 	 \jline
{\small rice}			& \ccol{0.0} 	& \ccol{1.1}  & \ccol{1.0} 	    & \ccol{5.0} 	 & \ccol{1.9} 	 & \ccol{0.7} 	 		& \ccol{2.3} 	 & \ccol{0.7}  & \ccol{1.7} 	& \ccol{0.0} 	& \ccol{0.0}  & \ccol{0.5} 	 & \ccol{1.6}  & \ccol{0.6} 	& \ccol{0.0}	& \ccol{3.4} 	 & \ccol{69.3} 		 & \ccol{0.0} 	 & \ccol{2.6} 	 	 	 & \ccol{0.0} 	  & \ccol{4.5}  & \ccol{0.5} 	 		& \ccol{0.6 } 	& \ccol{0.6} 	 	& \ccol{0.5} 	 \jline
{\small soda} 			& \ccol{0.0} 	& \ccol{0.0}  & \ccol{0.0} 	    & \ccol{0.0} 	 & \ccol{1.1} 	 & \ccol{0.9} 	 		& \ccol{1.5} 	 & \ccol{0.0}  & \ccol{0.0} 	& \ccol{0.4} 	& \ccol{1.6}  & \ccol{8.5} 	 & \ccol{0.0}  & \ccol{0.0} 	& \ccol{0.8}	& \ccol{0.0} 	 & \ccol{0.0} 		 & \ccol{71.1}	 & \ccol{0.0} 	 	 	 & \ccol{0.0} 	  & \ccol{0.0}  & \ccol{2.3} 	 		& \ccol{0.0 } 	& \ccol{3.2} 	 	& \ccol{8.2} 	 \jline
{\small spices} 		& \ccol{0.4} 	& \ccol{0.0}  & \ccol{0.4} 	    & \ccol{1.0} 	 & \ccol{0.5} 	 & \ccol{0.6} 	 		& \ccol{1.3} 	 & \ccol{0.0}  & \ccol{0.4} 	& \ccol{0.5} 	& \ccol{2.4}  & \ccol{2.1} 	 & \ccol{1.8}  & \ccol{0.4} 	& \ccol{0.6}	& \ccol{0.0} 	 & \ccol{0.0} 		 & \ccol{0.9} 	 & \ccol{80.1} 	 	 	 & \ccol{1.3} 	  & \ccol{0.8}  & \ccol{1.0} 	 		& \ccol{0.9 } 	& \ccol{0.8} 	 	& \ccol{1.0} 	 \jline
{\small sugar} 			& \ccol{0.7} 	& \ccol{0.7}  & \ccol{1.3} 	    & \ccol{0.0} 	 & \ccol{0.6} 	 & \ccol{0.0} 	 		& \ccol{0.7} 	 & \ccol{0.0}  & \ccol{2.3} 	& \ccol{0.0} 	& \ccol{0.7}  & \ccol{0.7} 	 & \ccol{2.8}  & \ccol{2.0} 	& \ccol{0.0}	& \ccol{0.0} 	 & \ccol{1.4} 		 & \ccol{0.0} 	 & \ccol{1.1} 	 	 	 & \ccol{82.0} 	  & \ccol{1.1}  & \ccol{1.1} 	 		& \ccol{0.0 } 	& \ccol{0.0} 	 	& \ccol{0.0} 	 \jline
{\small tea} 			& \ccol{0.0} 	& \ccol{0.0}  & \ccol{2.4} 	    & \ccol{2.5} 	 & \ccol{0.3} 	 & \ccol{3.2} 	 		& \ccol{2.9} 	 & \ccol{0.0}  & \ccol{0.0} 	& \ccol{0.0} 	& \ccol{0.3}  & \ccol{2.1} 	 & \ccol{0.7}  & \ccol{0.3} 	& \ccol{1.0}	& \ccol{0.8} 	 & \ccol{0.0} 		 & \ccol{0.3} 	 & \ccol{0.3} 	 	 	 & \ccol{0.3} 	  & \ccol{81.0} & \ccol{0.6} 	 		& \ccol{0.3 } 	& \ccol{0.0} 	 	& \ccol{0.0} 	 \jline
{\small tomato sauce} 	& \ccol{1.1} 	& \ccol{0.0}  & \ccol{0.6} 	    & \ccol{0.6} 	 & \ccol{0.5} 	 & \ccol{1.0} 	 		& \ccol{0.5} 	 & \ccol{0.0}  & \ccol{0.5} 	& \ccol{0.5} 	& \ccol{3.5}  & \ccol{1.8} 	 & \ccol{1.1}  & \ccol{0.6} 	& \ccol{0.0}	& \ccol{1.2} 	 & \ccol{0.0} 		 & \ccol{0.5} 	 & \ccol{1.8} 	 	 	 & \ccol{0.0} 	  & \ccol{0.0}  & \ccol{82.9} 	    	& \ccol{0.6 } 	& \ccol{0.0} 	 	& \ccol{0.0} 	 \jline
{\small tuna} 			& \ccol{0.0} 	& \ccol{0.0}  & \ccol{0.0} 	    & \ccol{0.9} 	 & \ccol{0.0} 	 & \ccol{3.1} 	 		& \ccol{1.6} 	 & \ccol{1.1}  & \ccol{0.0} 	& \ccol{0.7} 	& \ccol{0.0}  & \ccol{0.0} 	 & \ccol{0.7}  & \ccol{1.1} 	& \ccol{0.0}	& \ccol{0.7} 	 & \ccol{0.0} 		 & \ccol{0.0} 	 & \ccol{0.0} 	 	 	 & \ccol{0.0} 	  & \ccol{3.5}  & \ccol{0.0} 	 		& \ccol{85.3} 	& \ccol{0.0} 	 	& \ccol{0.8} 	 \jline
{\small vinegar} 		& \ccol{0.0} 	& \ccol{0.0}  & \ccol{0.5} 	    & \ccol{0.0} 	 & \ccol{0.0} 	 & \ccol{0.0} 	 		& \ccol{1.4} 	 & \ccol{0.0}  & \ccol{0.0} 	& \ccol{0.7} 	& \ccol{1.7}  & \ccol{9.7} 	 & \ccol{0.5}  & \ccol{0.0} 	& \ccol{6.3}	& \ccol{0.0} 	 & \ccol{0.0} 		 & \ccol{1.0} 	 & \ccol{1.8} 	 	 	 & \ccol{0.0} 	  & \ccol{0.7}  & \ccol{0.0} 	 		& \ccol{0.7 } 	& \ccol{73.6} 	 	& \ccol{0.7} 	 \jline
{\small water} 			& \ccol{0.0} 	& \ccol{0.0}  & \ccol{0.7} 	    & \ccol{0.4} 	 & \ccol{0.3} 	 & \ccol{0.4} 	 		& \ccol{0.3} 	 & \ccol{0.0}  & \ccol{0.0} 	& \ccol{0.0} 	& \ccol{0.3}  & \ccol{0.7} 	 & \ccol{0.0}  & \ccol{0.4} 	& \ccol{0.6}	& \ccol{0.0} 	 & \ccol{0.0} 		 & \ccol{1.0} 	 & \ccol{0.0} 	 	 	 & \ccol{0.0} 	  & \ccol{0.0}  & \ccol{0.8} 	 		& \ccol{0.0 } 	& \ccol{0.3} 	 	& \ccol{93.2} 	 \jline

\noalign{\global\let\w\undefined}%
\end{tabular}
\caption{The confusion matrix averaged over the five test splits.
        We achieve a mean accuracy of 78.9\% over all classes.}
\label{fig:confmat}
\end{figure*}

\input{figures/overview}

\end{document}